\documentclass[runningheads]{llncs}
\usepackage{booktabs}
\usepackage{multirow}
\usepackage{makecell}

\usepackage{amssymb}
\usepackage{amsmath}
\usepackage[T1]{fontenc}
\usepackage{graphicx,verbatim}
\usepackage{multirow}
\usepackage{url}

\usepackage{hyperref}
\usepackage[misc]{ifsym}
\usepackage{orcidlink}
\usepackage{amssymb}
\usepackage{booktabs} 
\usepackage{threeparttable}

\usepackage{color}
\hypersetup{
    colorlinks=true,
    linkcolor=blue,
    filecolor=blue,
    urlcolor=blue,
    citecolor=blue
}

\begin{document}
\title{ContiStain: Cross-Domain Relation-Preserving Distillation for Continual Multi-Domain Virtual IHC Staining}
\titlerunning{ContiStain}
%


\author{Fuqiang Chen  
\inst{1}\orcidlink{0009-0007-2345-1348}
\and Yifeng Wang 
\inst{2}
\and Hongpeng Wang  
\inst{1}
\and Yongbing Zhang  
\inst{1(\textrm{\Letter})}
} 

\authorrunning{F. Chen et al.}

\institute{School of Computer Science and Technology, Harbin Institute of Technology (Shenzhen), Shenzhen, China \\ \email{ybzhang08@hit.edu.cn}
\and School of Computer Science and Technology, Tsinghua University, Beijing, China
}
\maketitle              
\begin{abstract}


A unified multiplex virtual staining model enables scalable and non-destructive multiplex analysis from H\&E slides while promoting parameter efficiency, shared pathological knowledge, and consistent cross-biomarker representations.
However, in clinical practice, data for new biomarkers are typically acquired sequentially over time. Fine-tuning on such temporally arriving data leads to severe performance degradation on previously learned biomarkers, as sequential optimization disrupts the structured relationships among biomarker representations in the latent space.
To address this issue, we propose ContiStain, an IHC multi-domain relational distillation framework for continual virtual staining. We first (i) construct a domain-aware structured feature space using a mixture-of-experts (MoE) feature extractor to reduce representation interference across biomarker domains. Based on this stabilized feature space, we then (ii) 
propose a relation-preserving distillation strategy that explicitly enforces the consistency of cross-domain token-level cosine similarity matrices between learned biomarker domains during continual adaptation. 
By maintaining cross-domain structural coherence, ContiStain mitigates forgetting while retaining adaptability to new domains. Experiments on the MIST dataset under a four-domain sequential virtual IHC staining setting show improved stability, reducing FID and ConchFID by 11.1 and 60.9 compared to sequential fine-tuning, enabling scalable and robust multi-domain virtual staining. Code is released at \href{https://github.com/ccitachi/ContiStain}{https://github.com/ccitachi/ContiStain}.


\end{abstract}
\section{Introduction}

\begin{figure*}[tbp]
\centerline{\includegraphics[width=\textwidth]{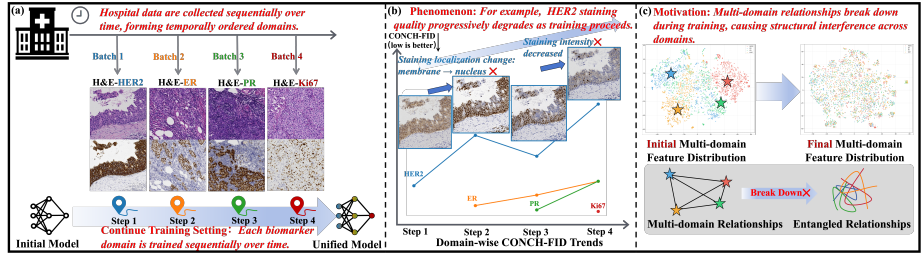}}
\caption{Challenges of multi-domain continual virtual staining. 
(a) Sequential arrival of multiple biomarkers. 
(b) Naïve fine-tuning causes catastrophic forgetting (CONCH-FID increases). 
(c) Cross-domain relationships gradually collapse, motivating relationship-aware learning.}

\label{intro}
\end{figure*}

With the growing clinical demand for efficient pathological analysis, virtual immunohistochemistry (IHC) staining has emerged as a practical approach to generate biomarker-specific images directly from routine H\&E slides \cite{klockner2025h}. Beyond single-biomarker translation \cite{chen2024pathological,peng2026usigan,guan2025supervised}, developing a unified multiplex virtual staining model \cite{lin2022unpaired,zhang2025pd,liu2025stainexpert,chen2026pgvms,xiong2025unpaired,dubey2024vims} capable of generating multiple biomarker domains within a shared framework is highly desirable for parameter efficiency, shared pathological knowledge, and consistent cross-biomarker representation.
In real-world pathology practice, however, new biomarkers are often adopted gradually based on evolving diagnostic needs and treatment decisions, rather than being introduced simultaneously. This naturally leads to a continual multi-domain virtual staining setting in Fig. \ref{intro} (a), where a unified model must incrementally adapt to new biomarker domains while preserving performance on previously learned ones.

In continual learning, sequential training often leads to the degradation of previously learned domains—in the case of virtual IHC staining, the generation quality of earlier biomarkers collapses in Fig. \ref{intro} (b).
Therefore, a number of methods have been proposed, primarily in the context of classification tasks. Replay-based approaches such as iCaRL~\cite{rebuffi2017icarl} mitigate forgetting by storing and revisiting a subset of samples from previous tasks. Distillation-based methods, including Learning without Forgetting (LwF)~\cite{li2017learning}, encourage consistency between the outputs of the 
old and updated models to preserve prior knowledge. Regularization-based strategies such as Elastic Weight Consolidation (EWC)~\cite{kirkpatrick2017overcoming} constrain important parameters to remain close to their previously learned values. These approaches have shown effectiveness in alleviating catastrophic forgetting in sequential supervised settings.

However, in continual multi-domain virtual IHC staining, preserving cross-biomarker structural relations is of central importance. Different IHC biomarkers exhibit distinct staining patterns—for example, HER2 primarily localizes to the cell membrane, whereas ER, PR, and Ki-67 are expressed in the cell nucleus. Despite these differences in molecular expression and spatial localization, all biomarkers are observed within the same underlying tissue architecture, sharing consistent glandular structures, cellular arrangements, and tumor regions. Consequently, their latent representations are inherently structured and interrelated. If these shared structural relations are disrupted during sequential adaptation, the model gradually loses its alignment to the underlying tissue morphology, resulting in representation drift and unstable generation as shown in Fig. \ref{intro} (c).


To address the above challenges, we propose an \textbf{IHC multi-domain relational distillation continual learning framework (ContiStain)}. To the best of our knowledge, this work is the first to formally establish continual multi-domain virtual IHC staining as a dedicated learning setting and to explicitly emphasize cross-biomarker structural preservation as a core objective.
To maintain stable cross-biomarker relations throughout sequential domain adaptation, ContiStain is built upon two progressively structured components.
First, we construct a domain-aware feature extractor using  a Mixture-of-Experts module \cite{shazeer2017outrageously}, 
where biomarker-specific experts model domain characteristics while shared components preserve morphology-aligned structures, 
providing a stable foundation for continual adaptation.
Second, we introduce a relational distillation strategy that preserves cross-domain structural consistency within the MoE-structured latent space established in the first step, ensuring coherent structural relations among learned biomarkers during sequential updates.
Extensive experiments demonstrate that ContiStain effectively mitigates forgetting while maintaining cross-domain stability in continual multi-domain virtual IHC staining. 

The main contributions are as follows:

\textbf{1)} We introduce a relational distillation mechanism that operates within the structured latent space to explicitly maintain cross-domain structural consistency across biomarkers.

\textbf{2)} We propose a domain-aware structured feature modeling strategy based on MoE, which disentangles biomarker-specific characteristics from morphology-aligned structures to support stable continual adaptation.

\textbf{3)}  Extensive experiments on the MIST dataset demonstrate the effect of the proposed framework under a sequential multi-domain setting.

\section{Method}

We consider a sequential multi-domain scenario in which 
biomarker domains $\{D_1, \dots, D_K\}$ are introduced progressively. 
At training step $t$, only data from the current domain $D_t$ are accessible, 
while previous data are unavailable.

After training on $D_{t-1}$, the generator is frozen as a teacher model $G^{t-1}$, serving as a stable reference. During adaptation to $D_t$, the student model $G^t$ is optimized using current-domain supervision together with a structural preservation objective computed against $G^{t-1}$ on previous domains. Unlike replay-based methods, we preserve knowledge via cross-biomarker structural relations.

\begin{figure*}[t]
\centerline{\includegraphics[width=\textwidth]{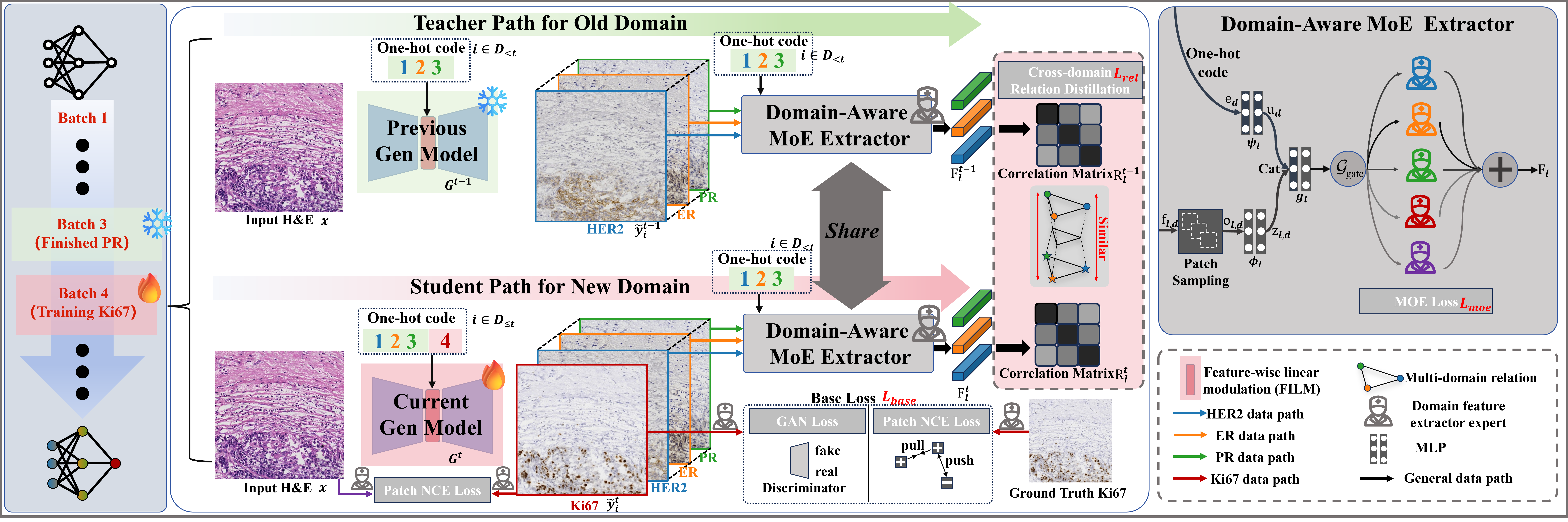}}
\caption{ContiStain: continual virtual staining via MoE-structured features and cross-domain relational distillation.  The fourth domain process is used for illustration.}
\label{method}
\end{figure*}

\subsection{Base Multi-domain Virtual IHC Framework}

We build our framework upon Adaptive Supervised PatchNCE (ASP), 
an encoder--decoder generative model for H\&E-to-IHC translation 
trained with adversarial and patch-level contrastive supervision.

Given an H\&E image $x$ and a biomarker domain index 
$d \in \{1,\dots,K\}$, the generator produces a stain-specific output:
\begin{equation}
\hat{y}_d = G(x, d)
\end{equation}
In our setting, domain indices correspond to different biomarkers 
(e.g., $1$: HER2, $2$: ER, $3$: PR, $4$: Ki67).

Bottleneck-only domain injection preserves domain-invariant morphology and confines biomarker-specific modulation to a compact latent space by applying feature-wise linear modulation (FiLM)~\cite{perez2018film} at the bottleneck, where the domain index is encoded as a one-hot vector $\mathbf{e}_d$ and mapped to an embedding $\mathbf{u}_d = \psi(\mathbf{e}_d)$.


Let $\mathbf{h}_b$ denote the bottleneck feature  following \cite{lin2022unpaired}.
FiLM modulates it as:

\begin{equation}
\mathbf{h}_b' =
(1 + \boldsymbol{\gamma}_b(d)) \odot \mathbf{h}_b
+ \boldsymbol{\beta}_b(d)
\end{equation}
where $[\boldsymbol{\gamma}_b(d), \boldsymbol{\beta}_b(d)] 
= \phi_b(\mathbf{u}_d)$.
This builds a lightweight unified conditional virtual IHC backbone for continual cross-domain relation preservation.

\subsection{Domain-Aware MoE Extractor for Relational Modeling}

A key challenge in continual multi-domain virtual staining is 
\emph{representation interference}: as the generator adapts to a new biomarker domain, 
the latent representations of previously learned domains are implicitly reshaped. 
Such interference destabilizes cross-domain structural relations, 
making relation preservation difficult to enforce.
To address this issue, we construct a \emph{domain-aware} feature extractor 
$F_{\mathrm{MoE}}$  that produces structured and disentangled representations 
for different biomarker domains.

\noindent{{\textbf{Patch Sampling and Shared Projection:}}}
For each biomarker domain $d \in \mathcal{D}$, 
we extract intermediate encoder features 
$\{\mathbf{f}_{\ell,d}\}_{\ell \in \mathcal{L}}$, where 
$\mathbf{f}_{\ell,d} \in \mathbb{R}^{B \times N_\ell \times C_\ell}$ 
denotes patch-level tokens at layer $\ell$ 
($N_\ell$ is the number of spatial tokens).

We randomly sample $P$ spatial tokens to obtain a compact set of structural descriptors, which are then projected into a shared embedding space:

\begin{equation}
\mathbf{o}_{\ell,d}
= \mathrm{Sample}(\mathbf{f}_{\ell,d}) 
\in \mathbb{R}^{B \times P \times C_\ell}, \quad
\mathbf{z}_{\ell,d} 
= \phi_{\ell}(\mathbf{o}_{\ell,d}) 
\in \mathbb{R}^{B \times P \times C_\ell}
\end{equation}
where $\phi_\ell$ is a two-layer MLP applied to each token independently. 
The projection function is shared across domains to align patch-level representations in a unified latent space. This shared space is designed to emphasize morphology-related structures rather than biomarker-specific appearance, enabling meaningful cross-domain relational comparison.

\noindent{{\textbf{Domain-Aware Mixture-of-Experts Modeling:}}}
To explicitly incorporate domain structure while preserving content awareness,
we introduce a domain-aware mixture-of-experts module.

The biomarker domain $d$ is encoded as a one-hot vector
$\mathbf{e}_d$ and mapped to a compact embedding $\mathbf{u}_d = \psi(\mathbf{e}_d)$.
Importantly, the expert routing is conditioned on both the
domain embedding and the image content representation. We denote the gating network by $\mathcal{G}_{\mathrm{gate}}$, which corresponds to the $\mathcal{G}_{\mathrm{gate}}$ block shown in Fig.~\ref{method}.
For each token embedding $\mathbf{z}_{\ell,d}$, the gating network $\mathcal{G}_{\mathrm{gate}}$ predicts domain- and content-adaptive expert weights:



\begin{equation}
\mathbf{w}_{\ell} = 
\mathrm{softmax}\Big(
g_\ell([\mathbf{z}_{\ell,d}, \mathbf{u}_d])
\Big)
\end{equation}
where $M$ experts $\{h_{\ell,m}\}_{m=1}^{M}$ produce the
domain-aware structured feature:

\begin{equation}
{\mathbf{F}}_{\ell,d}
=
\mathrm{Normalize}
\!\left(
\sum_{m=1}^{M}
\mathbf{w}_{\ell,m}
\odot
h_{\ell,m}(\mathbf{z}_{\ell,d})
\right)
\end{equation}

This domain-aware structured feature space reduces 
cross-domain interference and provides a stable basis 
for relation preservation.

\noindent{{\textbf{MoE Regularization.}}}
To enforce domain-consistent routing and avoid expert collapse, we regularize the gating probabilities, with the first term promoting domain-aligned expert preference and the second discouraging imbalanced expert usage:
\begin{equation}
\mathcal{L}_{\mathrm{MoE}}
=
\lambda_1 \mathrm{BCE}(\mathbf{w}_{\ell,m^\star}, y_d)
+
\lambda_2
\Big(
\max_m \bar{p}_m - \tfrac{1}{M}\sum_{m=1}^{M} \bar{p}_m
\Big)
\end{equation}
where $\mathbf{w}_{\ell,m^\star}$ denotes the routing probability of  reference expert $m$,
$y_d$ indicates the domain label, and $\bar{p}_m$ is the mini-batch average routing probability.

\subsection{Multi-Domain Relational Distillation}

While the domain-aware MoE builds a structured feature space,
continual updates can still \emph{distort cross-domain geometry}.
We thus distill \emph{relations} among previously learned biomarker domains
using a frozen teacher from the last step.

\noindent{{\textbf{Teacher--Student Generation for Old Domains.}}}
At continual step $t$, we keep the previous generator $G^{t-1}$ frozen as the teacher
(\texttt{preGAN}) and optimize the student $G^{t}$.
Given an input H\&E image $x$, for each old domain $i \in \mathcal{D}_{<t}$,
we generate old-domain outputs from both teacher and student:
\begin{equation}
\tilde{y}^{t-1}_i = G^{t-1}(x, i) ,
\qquad 
\tilde{y}^{t}_i = G^{t}(x, i)
\end{equation}

We then extract multi-layer encoder features 
and apply the same patch sampling and MoE projection module $F_{\mathrm{MoE}}$
to obtain structured feature tokens:
\begin{equation}
\mathbf{F}^{(\cdot)}_{\ell,i} 
= F_{\mathrm{MoE}}\!\left(E_{\ell}(\tilde{y}^{(\cdot)}_i),\, i\right)
\in \mathbb{R}^{P \times C}
\quad
(\cdot)\in\{t-1,t\},\ \ \ell\in\mathcal{S}.
\end{equation}
where $\mathcal{S}$ denotes the set of selected feature scales.

Importantly, the domain index $i$ and patch
 feature $E_{\ell}(\tilde{y}^{(\cdot)}_i)$ are explicitly provided to the MoE router
to ensure domain-aware structured features.

\noindent{{\textbf{Cross-Domain Relation Distillation.}}}
For each pair of old domains $(i,j)$, we compute a patch-level cosine correlation matrix
in the structured space:
\begin{equation}
\mathbf{R}^{(\cdot)}_{\ell}(i,j)
=
\mathrm{Normalize}(\mathbf{F}^{(\cdot)}_{\ell,i})
\;\cdot
\mathrm{Normalize}(\mathbf{F}^{(\cdot)}_{\ell,j})^\top
\in \mathbb{R}^{P \times P}
\end{equation}

We then preserve cross-domain geometry by minimizing the discrepancy between the teacher
and student relation matrices across selected layers and pairs:
\begin{equation}
\mathcal{L}_{\mathrm{rel}}
=
\frac{1}{|\mathcal{P}||\mathcal{S}|}
\sum_{(i,j)\in\mathcal{P}}
\sum_{\ell\in\mathcal{S}}
\left\|
\mathbf{R}^{t}_{\ell}(i,j)
-
\mathbf{R}^{t-1}_{\ell}(i,j)
\right\|_{1}
\end{equation}
In practice, we use output-level $\ell_1$ distillation when learning the second domain
(as cross-domain supervision is underdetermined with only one domain pair), and switch to the above relational distillation
from the third domain onward.

\subsection{Overall Objective}

At training step $t$, the student model $G^t$ is optimized 
using supervision from the current domain $D_t$ together 
with the relational preservation objective.

The overall loss is formulated as:
\begin{equation}
\mathcal{L}_{total} =
\mathcal{L}_{base}
+ \lambda_{rel} \mathcal{L}_{rel}
+ \lambda_{moe} \mathcal{L}_{moe}
\end{equation}
where $\mathcal{L}_{base}$ denotes the original ASP training objective 
(including adversarial and contrastive terms) on the current domain samples $\tilde{y}^{t}_{D_t}$ and corresponding label, 
$\mathcal{L}_{rel}$ is the relational distillation loss on the previous domains, 
and $\mathcal{L}_{moe}$ represents optional routing regularization 
on the MoE feature extractor.

\section{Experiments}

\noindent{{\textbf{Datasets:}}}
Experiments are conducted on the Multi-IHC Stain Translation (MIST) dataset~\cite{li2023adaptive}, which contains paired 1024$\times$1024 H\&E–IHC patches at 0.4661 $\mu$m per pixel (20$\times$) for four biomarkers: HER2, Ki67, ER, and PR. The training set includes 4,642, 4,361, 4,153, and 4,139 pairs for HER2, Ki67, ER, and PR, respectively, with 1,000 testing pairs for each biomarker.
The availability of multiple biomarker domains derived from whole-slide images enables systematic evaluation of continual multi-domain virtual IHC staining.

\noindent{{\textbf{Implementation Details:}}} 
We adopt ASP \cite{li2023adaptive} as the backbone for virtual staining. All continual learning baselines are implemented on the same ASP framework, differing only in the adopted continual learning strategy. All input H\&E–IHC patches are resized to $512 \times 512$.
In the continual multi-domain setting, biomarker domains are learned sequentially without accessing samples from previously learned domains. Unless otherwise specified, each domain is trained for 50 epochs with a batch size of 1 on a single NVIDIA RTX 3090 GPU. The learning rate is set to $1\times10^{-4}$ using the Adam optimizer.

\noindent{{\textbf{Evaluations:}}} We assess image quality using PSNR~\cite{avcibas2002statistical} and SSIM~\cite{wang2004image}, and distribution-level similarity via FID~\cite{heusel2017gans}. To better capture histopathology-specific features, we report CONCH-FID~\cite{chen2026pgvms} using a pathology-pretrained CONCH encoder~\cite{lu2024visual}, which captures pathology-aware semantic similarities through pathology image-text contrastive pretraining and provides a more clinically relevant assessment of staining quality, and DISTS~\cite{ding2020image} for structural and textural perceptual consistency.

\begin{table*}[t]
\centering
\small
\setlength{\tabcolsep}{4pt}
\renewcommand{\arraystretch}{1}
\caption{Domain-wise retention analysis (HER2$\rightarrow$ER$\rightarrow$PR$\rightarrow$Ki67). The \textit{Initial} row shows performance after the first-stage training, and the remaining rows report final results after continual learning. Best results are highlighted in \textbf{bold} (excluding \textit{Initial}). }
\label{tab:retention_full}

\resizebox{\textwidth}{!}{
\begin{tabular}{l|ccccc|ccccc}
\Xhline{1.5pt} 
\multirow{2}{*}{Method} &
\multicolumn{5}{c|}{HER2} &
\multicolumn{5}{c}{ER} \\
\cline{2-11}
& FID$\downarrow$ & ConchFID$\downarrow$ & PSNR$\uparrow$ & SSIM$\uparrow$ & DISTS$\downarrow$
& FID$\downarrow$ & ConchFID$\downarrow$ & PSNR$\uparrow$ & SSIM$\uparrow$ & DISTS$\downarrow$ \\
\hline

Initial
& 67.0973 & 88.8182 & 14.9335 & 0.1544 & 0.3045
& 41.2629 & 45.6739 & 13.3665 & 0.1368 & 0.2739 \\
\hline
Seq-FT 
& 80.9059 & 265.9206 & 13.7123 & 0.1323 & 0.2898
& 62.7876 & 98.2752 & 13.7309 & 0.1526 & 0.2797 \\

LWF 
& 84.7352 & 148.6154 & 14.1219 & 0.1286 & 0.2830
& 53.1903 & 62.7852 & \textbf{14.6164} & \textbf{0.1766 }& 0.2765 \\

EWC 
& 78.0549 & 197.3761 & 13.3334 & 0.1130 & 0.2828
& 69.1685 & 78.3096 & 13.1532 & 0.1336 & 0.2776 \\

iCaRL
& 76.6097 & 198.2103 & \textbf{14.8120} & \textbf{0.1644} & 0.2923
& {48.4756} & 59.1977 & 13.2304 & 0.1365 & 0.2729 \\

Ours
& \textbf{66.1568} & \textbf{111.9423} & 13.6269 & 0.1153 & \textbf{0.2798}
& \textbf{47.4210} & \textbf{47.1893} & 13.8484 & 0.1419 & \textbf{0.2677} \\

\hline

\multirow{2}{*}{Method} &
\multicolumn{5}{c|}{PR} &
\multicolumn{5}{c}{Ki67} \\
\cline{2-11}
& FID$\downarrow$ & ConchFID$\downarrow$ & PSNR$\uparrow$ & SSIM$\uparrow$ & DISTS$\downarrow$
& FID$\downarrow$ & ConchFID$\downarrow$ & PSNR$\uparrow$ & SSIM$\uparrow$ & DISTS$\downarrow$ \\
\hline

Initial
& 41.8015 & 36.3097 & 14.3173 & 0.1572 & 0.2668
& -- & -- & -- & -- & --\\
\hline
Seq-FT 
& 63.9035 & 98.4476 & 13.7745 & 0.1579 & 0.2853
& \textbf{39.3537} & \textbf{33.0343} & 14.4889 & 0.1615 & \textbf{0.2478} \\

LWF 
& 46.1695 & 47.8434 & 13.9698 & 0.1473 & 0.2648
& 43.2973 & 38.9575 & 14.5766 & 0.1590 & 0.2503 \\

EWC 
& 78.0882 & 104.4298 & 13.7501 & 0.1377 & 0.2737
& 64.4175 & 49.9091 & 13.1746 & 0.1316 & 0.2655 \\

iCaRL
& 51.3151 & 69.0636 & 13.5846 & 0.1381 & 0.2729
& 50.5395 & 60.1791 & 14.7017 & 0.1602 & 0.2619 \\

Ours
& \textbf{44.8295} & \textbf{38.1893} & \textbf{14.7375 }&\textbf{0.1599} & \textbf{0.2561}
& 44.3424 & 54.6830 & \textbf{14.9143} & \textbf{0.1739} & 0.2603 \\

\Xhline{1.5pt} 
\end{tabular}
}
\end{table*}

\begin{table*}[t]
\centering
\scriptsize
\setlength{\tabcolsep}{12pt}
\renewcommand{\arraystretch}{0.95}
\caption{Performance under reversed domain order (Ki67$\rightarrow$PR$\rightarrow$ER$\rightarrow$HER2). 
The \textit{Initial} row shows performance after the first-stage training, and the remaining rows report final results after continual learning. Results are averaged across all four domains to evaluate robustness to domain sequence variation.  Best results are highlighted in \textbf{bold} (excluding \textit{Initial}).}
\label{tab:reverse_order}

\resizebox{\textwidth}{!}{
\begin{tabular}{l|ccccc}
\Xhline{1.2pt}
Method 
& FID$\downarrow$ 
& ConchFID$\downarrow$ 
& PSNR$\uparrow$ 
& SSIM$\uparrow$ 
& DISTS$\downarrow$ \\
\hline

Initial 
& 48.3852 & 50.5980 & 14.5974 & 0.1564 & 0.2666 \\
\hline
Seq-FT
& 64.1188 & 161.7601 & 13.5295 & 0.1444 & 0.2887 \\

LWF 
& 56.4544 & 70.2778 & \textbf{14.8404} & \textbf{0.1694} & 0.2766 \\

EWC 
& 70.5058 & 193.4794 & 14.1416 & 0.1621 & 0.2923 \\

iCaRL
& 55.9826 & 81.6900 & 13.4792 & 0.1389 & 0.2776 \\

Ours
& \textbf{51.9917} & \textbf{63.9831} & 14.7410 & 0.1637 & \textbf{0.2711} \\

\Xhline{1.2pt}
\end{tabular}
}
\end{table*}

\begin{figure}[!htbp]
\centering
\includegraphics[width=1.0\textwidth]{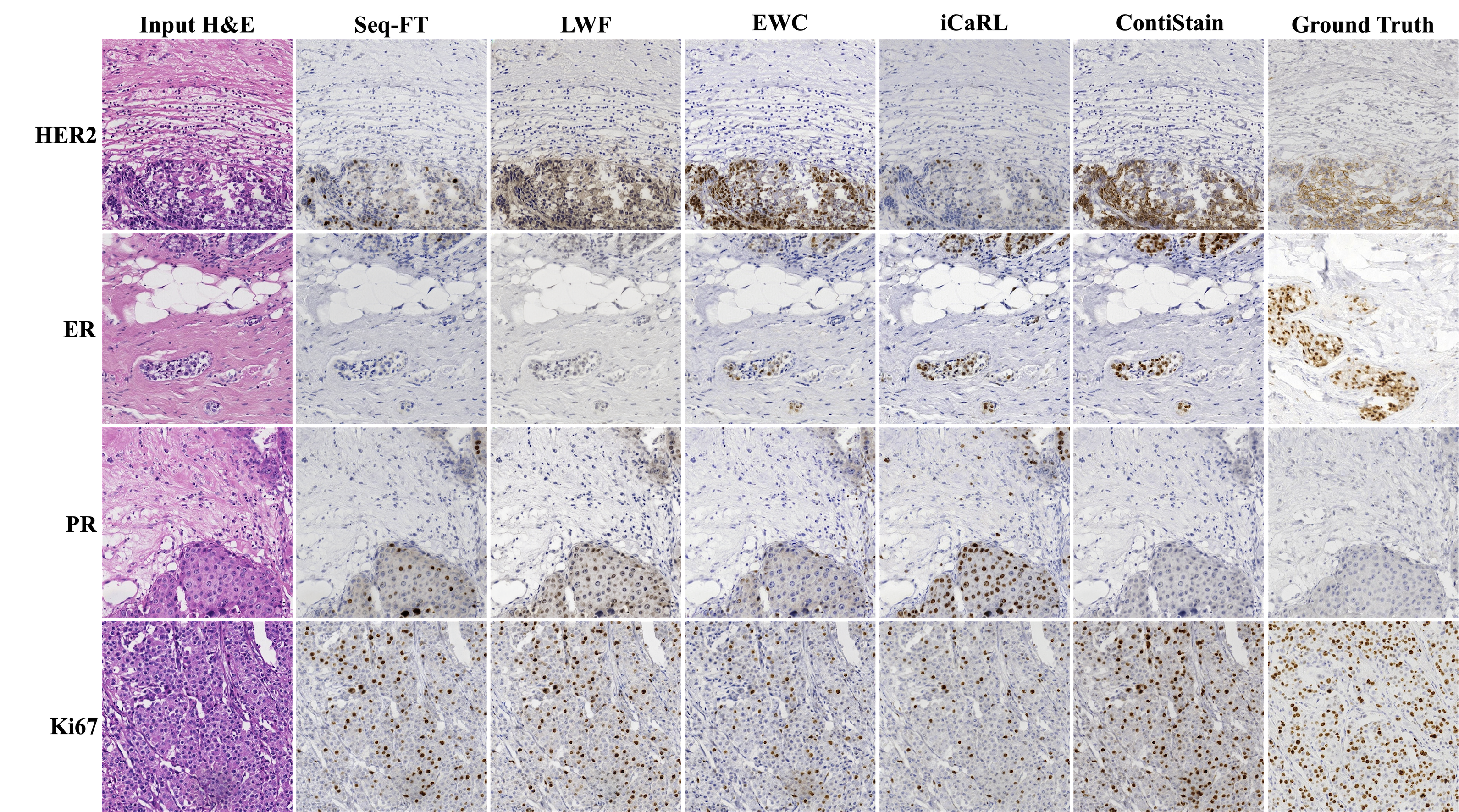}
\caption{Qualitative visualization of the final results on HER2, ER, PR, and Ki67.} \label{exp1}
\end{figure}
\begin{figure}[!htbp]
\includegraphics[width=\textwidth]{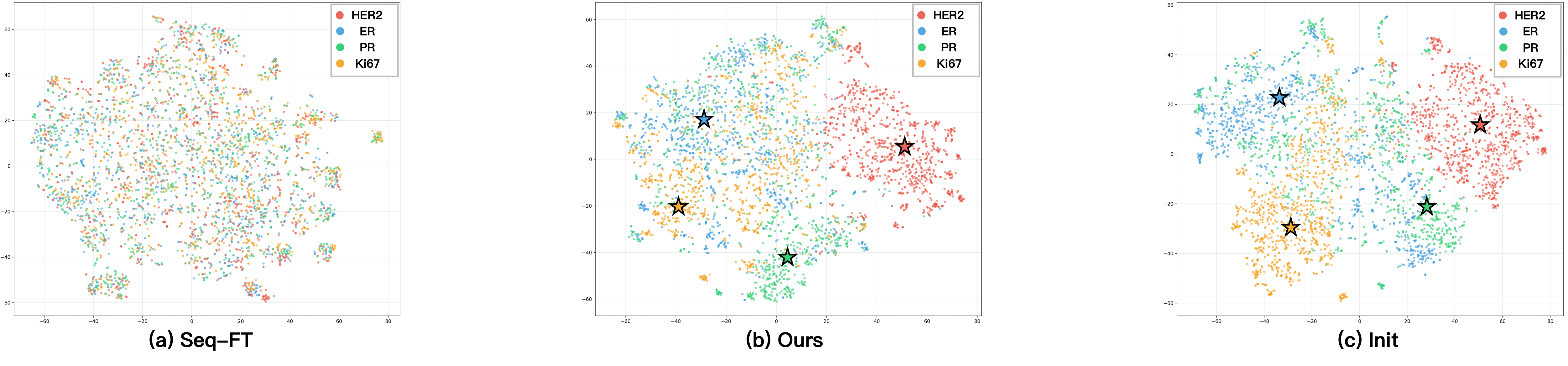}
\caption{Qualitative comparison of  distributions for Seq-FT(final), Ours(final), and Init.} \label{exp2}
\end{figure}

\noindent{\textbf{Comparison Study:}} Table~\ref{tab:retention_full} reports domain-wise retention after continual training (HER2$\rightarrow$ER$\rightarrow$PR$\rightarrow$Ki67). Our method achieves competitive or best performance, with lower FID/ConchFID on HER2/ER, superior PR results, and highest PSNR/SSIM on Ki67, indicating better pathological consistency and structural reconstruction. Table~\ref{tab:reverse_order} reports averaged results under the reversed domain order (Ki67$\rightarrow$PR$\rightarrow$ER$\rightarrow$HER2), indicating robustness to domain sequence variation. Fig.~\ref{exp1} illustrates that our method preserves HER2 membrane staining and achieves more accurate nuclear staining for ER, PR, and Ki67.  The t-SNE distributions in Fig.~\ref{exp2} show Seq-FT features scattered, while Ours aligns closely with Init, demonstrating improved cross-domain retention.

\noindent{\textbf{Ablation Study:}} 
Table~\ref{tab:ablation} presents ablation results on average performance.
Adding the relation distillation significantly improves FID and ConchFID.
Further incorporating MoE yields consistent gains across perceptual and fidelity metrics.
For the loss weight, $\lambda_{rel}=1$ achieves the best trade-off, while overly large weights degrade stability.
Regarding expert numbers, 5 experts (IHC and H\&E) provide the most balanced performance.

\begin{table*}[!htbp]
\centering
\scriptsize
\setlength{\tabcolsep}{8pt}
\renewcommand{\arraystretch}{1.0}
\caption{Ablation study on the average performance across all domains.}
\label{tab:ablation}

\resizebox{\textwidth}{!}{
\begin{tabular}{l l|ccccc}
\Xhline{1.2pt}
Type & Setting 
& FID$\downarrow$ 
& ConchFID$\downarrow$ 
& PSNR$\uparrow$ 
& SSIM$\uparrow$ 
& DISTS$\downarrow$ \\
\hline

\multirow{3}{*}{Module} 
& Baseline 
& 61.7377 & 123.9194 & 13.9267 & \textbf{0.1527} & 0.2757 \\
& + Rel Distill
& 57.9892 & 76.3321 & 13.8988 & 0.1454 & 0.2725 \\
& + Rel Distill+ MoE 
& \textbf{50.6874} & \textbf{63.0010} & \textbf{14.2818} & 0.1478 & \textbf{0.2660} \\

\hline
\multirow{4}{*}{Weight ($\lambda_{rel}$)} 
& 0.1 
& 57.1981 & 74.1813 & 14.2812 & 0.1497 & 0.2722 \\
& 1
& \textbf{50.6874} & \textbf{63.0010} & 14.2818 & 0.1478 & \textbf{0.2660} \\
& 10
& 89.8493 & 163.6526 & 14.6272 & \textbf{0.1628} & 0.2923 \\
& 100
& 121.3326 & 235.9494 & \textbf{14.7138} & 0.1614 & 0.3081 \\

\hline
\multirow{3}{*}{Experts} 
& 2 
& 65.8156 & 93.5733 & 14.2793 & 0.1451 & 0.2793 \\
& 5
& \textbf{50.6874} & \textbf{63.0010} & \textbf{14.2818} & \textbf{0.1478} & \textbf{0.2660} \\
& 10 
& 58.9102 & 86.5717 & 14.1437 & 0.1462 & 0.2762 \\

\Xhline{1.2pt}
\end{tabular}
}
\end{table*}

\section{Conclusion}

We formalize continual multi-domain virtual IHC staining and identify cross-biomarker structural preservation as its key challenge. To mitigate representation interference and structural drift, we propose ContiStain, a domain-aware relational distillation framework with structured MoE features and cross-domain structural consistency. Experiments on the four-domain MIST show it effectively reduces forgetting while maintaining structural coherence, suggesting that preserving latent cross-domain relations is a practical mechanism for scalable and robust continual virtual staining under real clinical constraints.

\bibliographystyle{splncs04}
\bibliography{Paper-0693}





\end{document}